%% file: main.tex
\documentclass{article}

\usepackage{microtype}
\usepackage{graphicx}
\usepackage{subcaption}
\usepackage{booktabs} 

\usepackage{hyperref}

\usepackage{ICML/arxiv2026}

\usepackage{amsmath}
\usepackage{amssymb}
\usepackage{mathtools}
\usepackage{amsthm}

\usepackage[capitalize,noabbrev]{cleveref}

\theoremstyle{plain}

\theoremstyle{definition}

\theoremstyle{remark}

\usepackage[textsize=tiny]{todonotes}

\icmltitlerunning{Practical FP4 Training for Large-Scale MoE Models on Hopper GPUs}

\begin{document}

\twocolumn[
  \icmltitle{Practical FP4 Training for Large-Scale MoE Models on Hopper GPUs}



  \icmlsetsymbol{equal}{*}

  \begin{icmlauthorlist}
    \icmlauthor{Wuyue Zhang}{equal,zjlab}
    \icmlauthor{Chongdong Huang}{equal,zjlab}
    \icmlauthor{Chunbo You}{zjlab}
    \icmlauthor{Cheng Gu}{zjlab}
    \icmlauthor{Fengjuan Wang}{zjlab}
    \icmlauthor{Mou Sun}{zjlab}
  \end{icmlauthorlist}

  \icmlaffiliation{zjlab}{Zhejiang Lab, Hangzhou, China}

  \icmlcorrespondingauthor{Mou Sun}{123sssmmm@gmail.com}


  \vskip 0.3in
]



\printAffiliationsAndNotice{}  

\begin{abstract}
Training large-scale Mixture-of-Experts (MoE) models is bottlenecked by activation memory and expert-parallel communication, yet FP4 training remains impractical on Hopper-class GPUs without native MXFP4 or NVFP4 support. In this work, we present a training recipe that enables MXFP4 efficiency for MoE models on Hopper architectures without native 4-bit computation support. A central challenge is to integrate FP4 into an existing BF16/FP8 hybrid training pipeline without incurring costly precision round-trips (e.g., FP4 $\leftrightarrow$ BF16 $\leftrightarrow$ FP8). We address this challenge by introducing direct FP8-to-FP4 quantization and de-quantization, together with scaling-aware FP4 row-wise to column-wise conversion, enabling FP4 activations and expert-parallel communication with minimal overhead. Core MoE computations are executed in FP8, while activations and expert-parallel communication are compressed using MXFP4, achieving substantial memory and bandwidth savings without degrading convergence. At the 671B parameter scale, our method achieves end-to-end training performance comparable to strong FP8 baselines, while reducing peak activation memory by 14.8\% (11.8 GB) and improving training throughput by 12.5\%, from 1157 to 1302 tokens per GPU per second. These results show that FP4 efficiency can be practically realized for large-scale MoE training through careful software-hardware co-design, even without native FP4 Tensor Core support. 
Codes are available at \href{https://github.com/anonymous-git-2026/MXFP4-Hopper}{github}.
\end{abstract}

\input{sections/01-introduction}
\input{sections/02-related-works}
\input{sections/03-methods}
\input{sections/04-experiments}
\input{sections/05-conclusion}



\bibliography{main}
\bibliographystyle{ICML/arxiv2026}

\newpage
\appendix
\onecolumn
\input{sections/appendix-a}


\end{document}

%% file: sections/01-introduction.tex
\section{Introduction}
\label{sec:intro}

The rapid scaling of Large Language Models (LLMs) into the 100B-1T parameter regime has pushed the limits of modern training infrastructure. To mitigate the growing demands on memory and computation, Mixture-of-Experts (MoE) architectures have emerged as a dominant paradigm for efficient scaling~\cite{gshard2021, mixtral2024, deepseekv3, kimik22025, gptoss2025}, activating only a sparse subset of expert parameters per token and thus decoupling model capacity from per-token compute.

Alongside architectural innovations, low-precision formats such as FP8 and FP4 have gained increasing attention for their ability to reduce memory footprint and improve arithmetic throughput, particularly for bandwidth- and memory-bound workloads. Prior work has shown that carefully designed low-precision training can preserve convergence while significantly improving system efficiency~\cite{deepseekv3, mxfp8recipe}. Recent hardware advances—most notably NVIDIA's Blackwell GPUs—enable native FP4 tensor core support, allowing aggressive quantization with minimal impact on convergence~\cite{mxfp4training2025, nvfp4training2025}. However, such support is not yet available on the majority of current-generation accelerators—particularly Hopper GPUs—which lack both FP4 compute and communication primitives.

This hardware gap presents significant challenges for adopting FP4 in existing training pipelines, especially for MoE models where activation memory and expert-parallel communication dominate system cost. Deploying FP4 on Hopper-class GPUs without native support introduces multiple technical obstacles: quantized formats such as MXFP4 rely on block-wise power-of-two scaling incompatible with Hopper's FP8 pipeline; intermediate conversions (e.g., FP8 $\leftrightarrow$ BF16 $\leftrightarrow$ FP4) incur latency and memory overhead; and communication libraries must be redesigned to support sub-byte packing and efficient dequantization.

In this work, we address these challenges by proposing a hybrid-precision MXFP4 training framework that enables FP4-level memory and communication efficiency on Hopper GPUs. Our approach compresses activations and all-to-all dispatches into software-emulated MXFP4 format, while preserving FP8 precision in compute-intensive GEMM operations. This design decouples compute and storage precision, maintains compatibility with Transformer Engine and DeepEP, and avoids lossy intermediate conversions—ensuring both numerical stability and practical deployment efficiency.

\vspace{1mm}
\noindent \textbf{Our contributions are as follows:}
\begin{itemize}
    \item We introduce an FP4 communication and caching strategy for expert-parallel MoE layers, reducing activation memory and inter-GPU traffic by over 50\%.
    \item We design a direct bitwise FP4-to-FP8 conversion algorithm with hierarchical scale alignment, eliminating the need for BF16 intermediates.
    \item We implement layout-aware CUDA kernels optimized for quantization, dispatch, and recomputation in FP4, including native support for ragged MoE tensors and transposed data layouts.
    \item We present the first production-scale deployment of software-emulated MXFP4 for MoE training on Hopper GPUs. At the 671B parameter scale, our method reduces peak activation memory by 14.8\% (11.8 GB), improves throughput by 12.5\% (from 1157 to 1302 tokens/GPU/s), and matches the convergence of strong FP8 baselines.
\end{itemize}

\begin{figure}[htb]
    \centering
    \includegraphics[width=\linewidth]{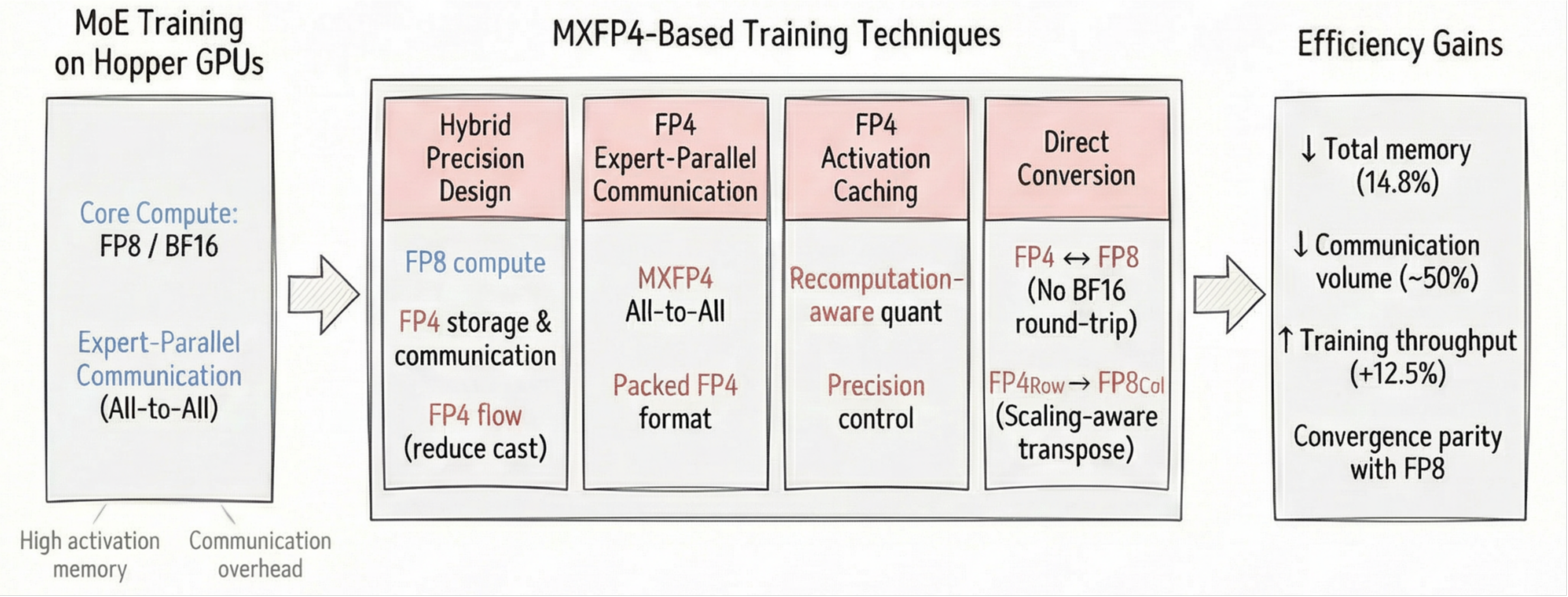}
    \caption{MXFP4-based MoE training on Hopper GPUs.}
    \label{fig:overview}
\end{figure}

We release our full training implementation (Appendix~\ref{sec:appendix-code}) to facilitate reproducibility and adoption. Figure~\ref{fig:overview} provides a high-level overview of our approach.

%% file: sections/02-related-works.tex
\section{Related works}
\label{sec:related}

Mixture-of-Experts (MoE) architectures have emerged as a dominant approach for scaling LLMs beyond 100B parameters~\cite{deepseekv3, kimik22025}. By activating only a sparse subset of expert layers per token, MoEs decouple model capacity from compute cost, enabling efficient training at trillion-parameter scales.

However, this efficiency comes at the cost of increased memory footprint and expert-parallel communication overhead. To further reduce training cost, recent efforts have turned to low-precision computation~\cite{peng2023fp8}. Formats such as FP8 and FP4 not only improve compute throughput—NVIDIA GB200 NVL72, for example, supports up to 1440 PFLOPS of FP4 tensor operations compared to 720 PFLOPS for FP8 and 360 PFLOPS for BF16~\cite{gb200_specs}—but also reduce activation memory, enabling larger batch sizes and model scales within the same hardware footprint. When combined with systems-level communication frameworks like DeepEP~\cite{deepep2025}, which support expert-parallel data transfer in FP8 formats, low-precision has become a standard technique for efficient training of large language models.

DeepSeek-V3~\cite{deepseekv3} demonstrated the first successful deployment of blockwise-scaled FP8 training at the 671B MoE scale on Hopper GPUs, using its custom FP8 kernel library DeepGEMM~\cite{deepgemm2025} to achieve high arithmetic throughput. This work validated FP8 as a viable alternative to BF16 in production-scale training, delivering substantial improvements in compute efficiency and throughput. In parallel, NVIDIA introduced MXFP8, a microscaled FP8 format with native tensor core support on Blackwell GPUs. Training recipes based on MXFP8~\cite{mxfp8recipe} combine power-of-two block-level scaling with stochastic rounding to match BF16 convergence while significantly reducing activation and weight footprint. Subsequent work~\cite{fp8flowmoe2025} shows that excessive precision casts in BF16-dominated pipelines can erode FP8's expected gains, motivating careful end-to-end dataflow design.

To further improve training efficiency, recent work has explored 4-bit floating-point formats such as MXFP4 and NVFP4. MXFP4 is a standardized narrow-precision floating-point format defined in~\cite{ocp_mx_specification}, which introduces low-bit formats such as MXFP8, MXFP6, and MXFP4 for efficient AI training and inference. Tseng et al.~\cite{mxfp4training2025} demonstrated that MXFP4 can be used effectively in the backward pass by combining stochastic rounding (SR) with randomized Hadamard transforms (RHT), achieving near-lossless convergence on 6.7B-scale models. Meanwhile, NVFP4 introduces smaller block sizes and E4M3-style scaling, and has been used to train 12B models with near-FP8 accuracy using two-level quantization and format-aware rounding heuristics~\cite{nvfp4training2025}. Chmiel et al.~\cite{chmiel2025fp4wayfullyquantized} further demonstrated that fully quantized end-to-end FP4 training of LLMs is feasible with carefully designed scaling and optimization strategies.

These advances suggest that FP4 formats can substantially improve training efficiency, but they typically rely on hardware with native FP4 tensor core support and have not yet been validated at larger MoE scales. While some large-scale models such as GPT-OSS-120B~\cite{gptoss2025} are rumored to employ FP4 training, no public documentation is available regarding their training configurations.

\paragraph{Motivation}
Unlocking FP4-level efficiency without native hardware support would immediately benefit ongoing workloads—particularly Mixture-of-Experts (MoE) models, where expert-parallelism amplifies both activation memory and all-to-all communication costs. If software-based MXFP4 quantization could be integrated into existing Hopper-compatible pipelines without degrading accuracy or throughput, the potential payoff would be considerable: reduced memory footprint for larger batch sizes, lighter communication payloads across GPUs, and more aggressive activation checkpointing.

However, realizing this in practice introduces multiple technical challenges. Hopper's FP8-centric transformer kernels expect specific layout and scale formats, making naive insertion of FP4 quantization difficult. Intermediate data type conversions (e.g., FP8 $\leftrightarrow$ BF16 $\leftrightarrow$ FP4) incur latency, precision loss, and memory overhead. Prior work on FP8-based MoE training~\cite{fp8flowmoe2025} highlights that excessive precision casts can significantly erode the efficiency gains of low-precision operators, underscoring the need for careful dataflow and operator design when integrating lower-precision formats into existing pipelines. Additionally, MXFP4 employs per-block power-of-two scaling and E2M1 encoding, which are structurally incompatible with Hopper's FP8 E4M3 layout, requiring careful bit-level remapping and scaling alignment. The recomputation strategy must also be aware of asymmetric precision across forward and backward passes to avoid instability.

This work is driven by the need to address these system-level obstacles and fully realize MXFP4's potential on existing Hopper infrastructure. Our approach introduces a hybrid precision framework that compresses activations and all-to-all communication into MXFP4 format, while retaining FP8 for compute-intensive paths. By decoupling storage and arithmetic precision, and co-designing the quantization, dispatch, and dequantization flows, we achieve substantial efficiency gains without relying on hardware changes.

%% file: sections/03-methods.tex
\section{Methodology}
\label{sec:methodology}

\begin{figure*}[t]
  \centering
  \includegraphics[scale=0.1]{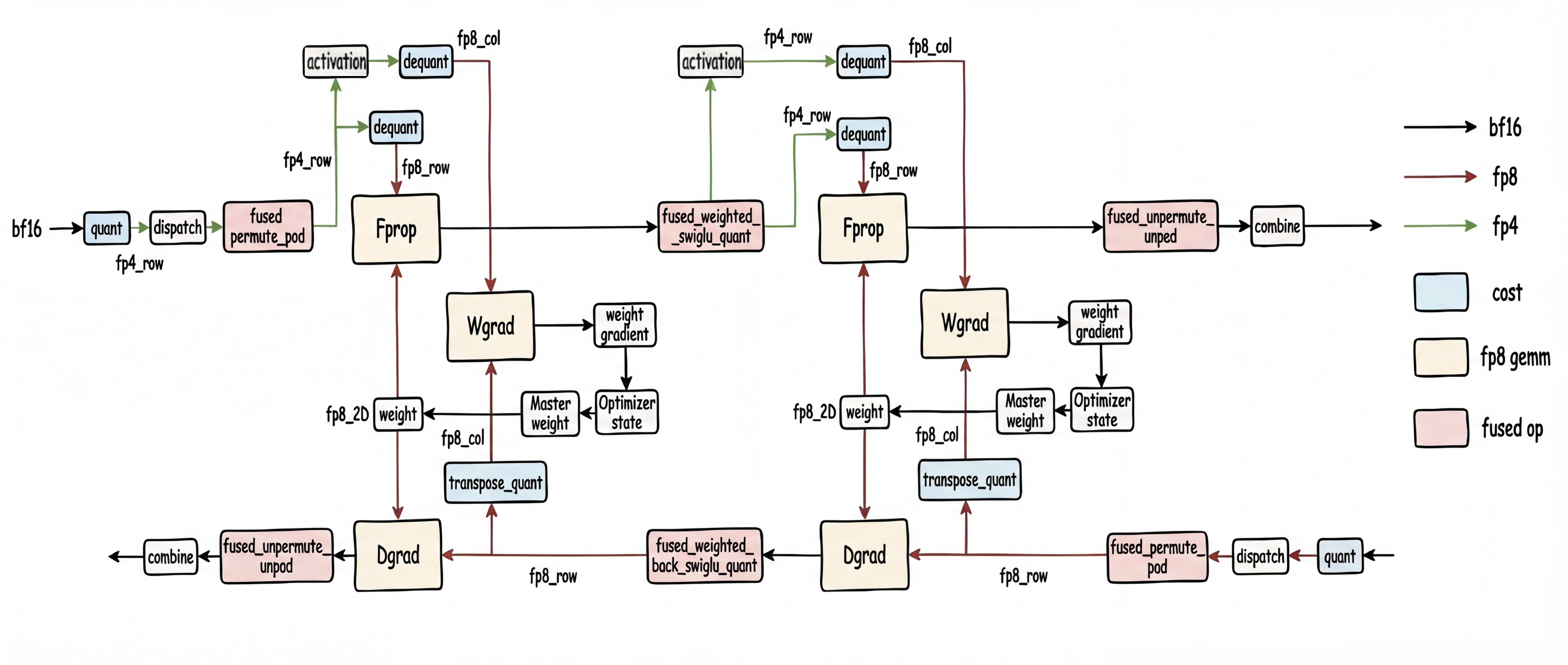}
    \caption{Overview of the MXFP4 hybrid-precision training flow.}
  \label{fp4flow}
\end{figure*}

We propose a soft-coded MXFP4 training framework as a plug-in extension to Transformer Engine’s blockwise FP8 recipe. FP4 elements are stored via bit-level packing into \texttt{uint8} containers. As illustrated in Figure~\ref{fp4flow}, our approach introduces a hybrid-precision dataflow designed to balance throughput with numerical stability.

Specifically, during the forward pass, activations are quantized into MXFP4 immediately before the All-to-All (A2A) dispatch to reduce communication volume, while the combine phase retains BF16 accumulation to preserve reduction accuracy. In addition, activation replicas required for recomputation are cached in MXFP4 format, significantly reducing the activation memory footprint.

In the backward pass, we intentionally revert to the standard FP8 communication and storage flow. Empirical profiling shows that the additional dequantization overhead introduced by FP4 (e.g., FP4 $\rightarrow$ FP8) outweighs potential communication savings for gradients. This asymmetric design—aggressive FP4 usage in the forward pass and conservative precision in the backward pass—achieves favorable end-to-end performance without compromising convergence.

To realize this flow efficiently, we develop specialized kernels---\texttt{BF16ToFP4Row}, \texttt{FP4RowToFP8Row}, and \texttt{FP4RowToFP8Col}---that perform layout-aware conversions bridging FP4 storage and FP8 computation.

\paragraph{Notation.}
Given FP4-quantized tensors with 4-bit values $X_{\mathrm{fp4}}$ and scaling factors $\mathrm{SF}_{\mathrm{fp4}}$, our conversion algorithm proceeds as follows. Hereafter, we use \textbf{SF} to denote scaling factor, \textbf{sign} for the sign bit, \textbf{exp} for exponent, and \textbf{mant} for mantissa. For clarity, scaling factors and sign bits are denoted using distinct symbols throughout. We use C-style bitwise operators throughout the paper.

\subsection{MXFP4 Quantization Specification on Non-Native Hardware}

In MXFP4, each block of 32 elements shares a common power-of-two scale factor and uses a 4-bit floating-point encoding with one sign bit, two exponent bits, and one mantissa bit (E2M1). This block-wise microscaling quantization achieves finer granularity than coarse per-tensor or per-row scaling schemes and has been shown to improve quantization fidelity and dynamic range in practice~\cite{ocp_mx_specification, micro_quant_arxiv}.

The underlying floating-point representation in MXFP4 follows the sign–exponent–mantissa convention of IEEE 754 formats. A real value is typically encoded with an implicit leading mantissa bit when the exponent is above its minimum; when the exponent bits are all zeros, values enter the subnormal (denormal) regime, where the implicit leading bit is dropped and the mantissa directly encodes fractional magnitude. Subnormal numbers extend dynamic range at the cost of leading precision, consistent with IEEE 754 definitions for small exponents~\cite{ieee754_floating_point}.

In our software implementation, we prioritize the computation of scaling factors. This design enables asynchronous scale write-backs to overlap with subsequent compute kernels, effectively hiding latency. The scale computation proceeds as follows.

For each $1 \times 32$ block of the input tensor, we determine the maximum absolute value, denoted as $\mathrm{group\_max}$. The floating-point scale is obtained by normalizing this value with respect to the maximum representable integer in MXFP4, denoted as $\mathrm{FP4\_MAX}=6$. Specifically, the intermediate scale is computed as $\mathrm{group\_max} / 6$ and then cast into the UE8M0 format by rounding up to the nearest power of two. Formally, the scaling factor is defined as
\[
\mathrm{SF} = 2^{\left\lceil \log_2 \left( \max(|X_{\mathrm{BF16}}|) / \mathrm{FP4\_MAX} \right) \right\rceil} .
\]
Once the scale factor is computed, quantization proceeds in two stages. First, the input values are numerically mapped and cast from BF16 to the MXFP4 representation. Second, the resulting low-precision values are reorganized via bit-level extraction and packing into \texttt{uint8} containers, ensuring compatibility with the NVIDIA Hopper memory layout and tensor core requirements.

The conversion begins by normalizing each BF16 input value using the corresponding floating-point scale, mapping it into the MXFP4 dynamic range. CUDA intrinsic functions are used to facilitate the casting from floating-point to integer representations. A key consideration in this process is the treatment of subnormal values. When the exponent falls below the minimum representable threshold, the implicit leading bit must be explicitly materialized and subjected to appropriate bit shifting. This mechanism trades mantissa precision for extended dynamic range, implementing gradual underflow.

The kernel implementation relies on two core bitwise operations. First, \emph{materializing the implicit bit}: following the IEEE~754 convention, normalized values contain an implicit leading bit of one. We apply the transformation $(\texttt{0x400000u} \,\texttt{|}\, (\mathrm{mant} \texttt{>>} 1))$, which right-shifts the 23-bit mantissa by one and performs a bitwise OR with \texttt{0x400000u}, explicitly reinserting the hidden bit into the most significant position of the fraction. Second, \emph{exponent alignment via right shifting}: the mantissa is further right-shifted by a computed offset, effectively increasing the exponent until it aligns with the minimum representable range of the target format.

Rounding behavior is another critical aspect of the implementation. For MXFP4, we adopt round-to-nearest, ties-to-even (RTNE). When truncating high-precision values (e.g., FP32 with a 23-bit mantissa) to ultra-low precision formats, naive truncation introduces systematic bias. RTNE, as the default rounding mode in IEEE~754, ensures that quantization error remains statistically zero-centered.

The rounding decision is derived from three signals: the last bit ($L$), the guard bit ($G$), and the sticky bit ($S$). Here, $L$ denotes the least significant bit of the target format (bit~22), where $L=1$ indicates an odd value and $L=0$ an even value. The guard bit $G$ corresponds to the first bit beyond the target precision (bit~21) and indicates whether the remainder is at least $0.5$. The sticky bit $S$ is defined as the logical OR of all remaining lower-order bits (bits~20--0). The final rounding increment is determined by
$R = G \,\&\, (S \,|\, L)$.

After consolidating the sign, exponent, and mantissa fields of the MXFP4 representation, two FP4 values are packed into a single \texttt{uint8} element. This final step consists of standard bit-manipulation and packing operations, which we omit for brevity.

\subsection{Loss-Neutral FP4-Enabled Mixed Precision Training Flow}

A key design choice is to place quantization immediately before the All-to-All (A2A) dispatch. FP4-quantized activations are packed into \texttt{uint8} containers, reducing both communication volume and memory footprint without altering the core computation logic. As a result, the framework remains fully compatible with existing FP8-based hardware acceleration pipelines.

\textbf{Quantization Timing and Communication Optimization.}
Strictly positioning quantization prior to A2A dispatch enables a substantial reduction in communication payload. Specifically, the standard FP8 communication format—consisting of FP8 activations with FP32 scaling factors—is replaced by a compact representation that combines packed FP4 activations with 8-bit integer scales.

Assuming an activation tensor of shape $M \times K$, the communication footprint can be analyzed as follows. In the FP8 baseline, activations are transmitted as \texttt{uint8} values, incurring $M \cdot K$ bytes, while scaling factors are stored as FP32 with a block size of 128, contributing an additional $(M \cdot K / 128) \cdot 4$ bytes. This results in a total payload of approximately $1.03 \cdot M \cdot K$ bytes. In contrast, under MXFP4, two FP4 values are packed into a single \texttt{uint8} byte, reducing the activation payload to $(M \cdot K)/2$ bytes. Scaling factors follow the MXFP4 specification and are stored as 8-bit integers with a block size of 32, contributing $(M \cdot K / 32) \cdot 1$ byte. The total communication volume is therefore approximately $0.53 \cdot M \cdot K$ bytes, corresponding to a reduction of roughly 50\%.

We implement this customized data flow within the DeepEP communication library by extending the A2A operator to support packed FP4 formats. Throughout this process, the quantization and de-quantization steps are designed to remain logically lossless with respect to the original FP8 computation.

\textbf{Decoupling Computation and Storage Precision.}
To balance computational throughput and memory efficiency, we adopt a decoupled precision strategy. All forward and backward GEMM operations are executed in FP8, allowing us to fully leverage native Tensor Core acceleration and preserve numerical stability. In contrast, activation tensors produced during the forward pass are immediately quantized and cached in HBM as MXFP4 replicas. Compared to FP8 activation caching, this strategy reduces activation memory usage by nearly 50\%, enabling larger batch sizes, more aggressive recomputation strategies, or further model scaling.

\textbf{Specialized Kernel Support.}
Supporting this hybrid-precision flow requires layout-aware format conversion. We therefore develop three specialized CUDA kernels. \texttt{BF16ToFP4Row} performs row-wise MXFP4 quantization and packing during the forward pass. \texttt{FP4RowToFP8Row} efficiently de-quantizes packed FP4 activations into FP8 row-major format for standard GEMM execution. Finally, \texttt{FP4RowToFP8Col} is designed for weight-gradient (Wgrad) computation, where a transposed view of the activation tensor is required. This kernel fuses de-quantization with matrix transposition, directly converting FP4 row-major storage into FP8 column-major format to minimize memory traffic.

\textbf{Strategic Fallback in Backward Propagation.}
Although FP4 provides significant storage and communication benefits, applying it uniformly across all training stages introduces trade-offs. Our profiling shows that, in the backward pass, the overhead of additional de-quantization steps (e.g., BF16 $\rightarrow$ FP4 $\rightarrow$ FP8) outweighs the potential communication savings for gradients. Consequently, we intentionally revert to a standard FP8 communication flow during backpropagation. This asymmetric design—aggressive FP4 usage in the forward pass and conservative precision in the backward pass—proves effective in optimizing end-to-end training throughput without compromising convergence.

\subsection{Direct FP4-to-FP8 Conversion Algorithm}

While NVIDIA Hopper GPUs provide efficient hardware support for FP8 matrix multiplication, they lack native instructions for FP4 computation. As a result, the efficiency of our approach critically depends on an optimized conversion pipeline that bridges storage-efficient MXFP4 representations and compute-efficient FP8 operands.

A direct conversion is non-trivial due to the structural mismatch between MXFP4 and FP8 formats. MXFP4 uses an \texttt{E2M1} encoding packed into \texttt{uint8} containers, with a block size of 32 and scaling factors stored in UE8M0 format. In contrast, FP8 adopts an \texttt{E4M3} layout with a block size of 128 and FP32 scaling. A naive conversion path that up-casts MXFP4 to BF16 and subsequently down-casts to FP8 (FP4 $\rightarrow$ BF16 $\rightarrow$ FP8) incurs substantial latency and memory overhead.

To avoid these costs, we propose a direct bit-wise FP4-to-FP8 conversion algorithm that bypasses the BF16 intermediate. The algorithm combines integer-domain format remapping with a hierarchical scaling alignment strategy to reconcile differences in mantissa layout, exponent bias, and block granularity. Algorithm~\ref{alg:DirectConversion} summarizes the complete FP4-to-FP8 conversion procedure.

\textbf{Bit-wise Format Remapping.}
The conversion begins by unpacking the sign, exponent, and mantissa fields from the FP4 bitstream. The exponent is adjusted directly in the integer domain according to
\[
E_{\text{fp8}} = E_{\text{fp4}} - \mathrm{Bias}_{\text{fp4}} + \mathrm{Bias}_{\text{fp8}}.
\]
The single-bit mantissa of the FP4 \texttt{E2M1} format is then left-shifted to align with the most significant bit of the FP8 \texttt{E4M3} mantissa field. This operation preserves relative precision while avoiding floating-point arithmetic during format conversion.

\textbf{Hierarchical Scaling Alignment.}
The primary challenge arises from the mismatch in block granularity: four MXFP4 blocks of 32 elements must be mapped onto a single FP8 block of 128 elements. We address this using a max-based hierarchical alignment strategy.

First, a target scale $S_{\text{target}}$ is selected as the maximum scaling factor among the four source MXFP4 blocks. Second, to preserve numerical equivalence, each element’s exponent is compensated by the difference between the local block scale and the target scale,
\[
E'_{\text{fp8}} = E_{\text{fp8}} - (S_{\text{target}} - S_{\text{sub}}).
\]
To mitigate underflow when the disparity between local and global scales is large, we introduce a small empirically chosen shift constant $\delta$, which is folded into the target scale selection during implementation.

Finally, because MXFP4 scaling factors are stored in UE8M0 format and represent exact powers of two, we map each 8-bit scale directly into the exponent field of an FP32 value via bit shifting. This avoids costly floating-point exponentiation and ensures hardware-efficient casting.

\begin{algorithm}[!hbtp]
    \caption{Direct FP4-to-FP8 Conversion Algorithm}
    \label{alg:DirectConversion}
    \renewcommand{\algorithmicrequire}{\textbf{Input:}}
    \renewcommand{\algorithmicensure}{\textbf{Output:}}
    
    \begin{algorithmic}[1]
        \REQUIRE $X_{\text{fp4}} (X_0 \sim X_3)$, $\mathrm{SF}_{\text{fp4}} (\mathrm{SF}_0 \sim \mathrm{SF}_3)$
        \ENSURE $X_{\text{fp8}}$, $\mathrm{SF}_{\text{fp8}}$
        
        \STATE $\mathrm{SF}_{\text{fp8}} = \max(\mathrm{SF}_{\text{fp4}}) - 6$
        \STATE $E_{i,\text{adjust}} = \mathrm{SF}_{\text{fp8}} - \mathrm{SF}_i$
        \STATE $\text{sign}_i = (X_i \texttt{>>} 3) \ \&\ 1$
        \STATE $E_{i,\text{fp4}} = (X_i \texttt{>>} 1) \ \&\ 3$
        \STATE $M_i = X_i \ \&\ 1$
        \STATE $E_{i,\text{fp8}} = E_{i,\text{fp4}} - \mathrm{Bias}_{\text{fp4}} + \mathrm{Bias}_{\text{fp8}}$
        \STATE $E_i = E_{i,\text{fp8}} - E_{i,\text{adjust}}$
        \STATE $X_{i,\text{fp8}} = (\text{sign}_i \texttt{<<} 7) \mid (E_i \texttt{<<} 3) \mid (M_i \texttt{<<} 2)$
        \STATE $X_{\text{fp8}} = \{X_{0,\text{fp8}} \sim X_{3,\text{fp8}}\}$
        
        \RETURN $X_{\text{fp8}}$, $\mathrm{SF}_{\text{fp8}}$
    \end{algorithmic}
\end{algorithm}

\subsection{High-Performance Kernel Implementation}

We implement a suite of highly optimized CUDA kernels for the FP4-enabled training pipeline. In particular, for weight gradient computation (Wgrad), which requires a transposed view of the activation tensor, we design a fused \texttt{FP4RowToFP8Col} kernel that integrates multiple optimizations to minimize memory traffic and conversion overhead.

\textbf{Operator fusion and layout transformation.}
Conventional implementations perform format conversion and memory layout transformation (i.e., transposition) as separate steps, incurring multiple global memory reads and writes. Our kernel instead fuses de-quantization, matrix transposition, and re-quantization into a single execution unit. By carrying out the layout transformation entirely within on-chip shared memory, we eliminate intermediate global memory accesses and reduce overall global memory bandwidth consumption by approximately $3\times$ compared to BF16-based baselines.

\textbf{Conflict-free shared memory access.}
To convert row-major FP4 storage into column-major FP8 output, we use shared memory as a transposition buffer. To avoid bank conflicts during column-wise accesses, we introduce padding in the shared memory tile, allocating it with a stride of \texttt{[TILE\_DIM][TILE\_DIM + 1]}. This skewed layout ensures that column-wise accesses map to distinct memory banks, enabling full utilization of the shared memory bandwidth.

\textbf{Bit-level arithmetic optimization.}
We exploit the bit-level structure of the MXFP4 and FP8 formats to avoid costly floating-point operations whenever possible: (i) \emph{Scale decoding.} MXFP4 scaling factors stored in UE8M0 format are decoded directly into BF16 by bit shifting (e.g., \texttt{val \texttt{<<} 7}), avoiding floating-point multiplication. (ii) \emph{Power-of-two alignment.} To conform with the scaling conventions used in Transformer Engine, we implement a bit-wise scaling procedure. Instead of computing $2^{\lceil \log_2(x) \rceil}$ via math library calls, we directly manipulate the exponent field of the IEEE~754 representation, significantly reducing instruction latency.

\textbf{Format-compatible expert parallel communication.} 
To support MXFP4 scales stored in compact integer representations (e.g., UE8M0), we generalize scale handling in the expert dispatch path to be byte-addressable and precision-agnostic. Specifically, we propagate the scale element size across intranode and internode communication, RDMA/NVL buffer layout, and kernel copy logic, so that scale payloads are treated as variable-sized byte regions rather than implicitly assumed FP32 tensors. This generalization is reflected in per-token communication footprint computation, clean-metadata derivation, and symmetric buffer offset calculation, eliminating buffer over-allocation and misalignment when integer-encoded scales are used. At the kernel level, scale transfers dynamically select between FP32 and byte-wise copy paths based on the propagated element size, while preserving alignment constraints required by TMA and warp-level copy mechanisms.

\textbf{Native support for ragged MoE tensors.}
In Mixture-of-Experts architectures, the number of tokens routed to each expert varies dynamically, resulting in ragged tensors. Our kernel natively supports variable-length splits through precomputed offsets (\texttt{split\_lens}, \texttt{split\_offsets}), dynamically computing output addresses for contiguous storage without padding. Vectorized stores are employed even on misaligned boundaries, with edge cases handled entirely within registers.

See Appendix~\ref{sec:appendix-code} for implementation details and code integration instructions.

%% file: sections/04-experiments.tex
\section{Experiments}
\label{sec:experiments}

In this section, we evaluate the system performance of the proposed MXFP4 training framework, focusing on training throughput, memory footprint, and recomputation efficiency.

\subsection{Experimental setup}
\textbf{Hardware \& Environment}. We conduct all experiments on a cluster of 32 nodes, equipped with a total of 256 NVIDIA Hopper GPUs (80GB HBM3). The GPUs are interconnected via NVSwitch within nodes and InfiniBand across nodes.

\textbf{Model Configuration}. We train a 671B-parameter Mixture-of-Experts (MoE) model featuring the Multi-Head Latent Attention (MLA) architecture. The model configuration follows the public DeepSeek-V3 settings, including the Multi-Head Latent Attention (MLA) architecture, with identical MoE expert and projection-layer dimensions.

\textbf{Baselines}. We compare our system against two strong baselines implemented in Megatron-LM:\\
1. BF16: Standard bfloat16 mixed-precision training.\\
2. FP8: Transformer engine blockwise FP8 recipe  training optimized for Hopper architecture. 

We report the training throughput in Tokens per GPU per Second (TGS) and peak memory usage.
 
\subsection{End-to-End Training Performance}

Table~\ref{tab:main_results} summarizes the performance comparison. Our primary observation is that memory constraints on the 671B model force standard baselines into aggressive recomputation (rematerialization) regimes, limiting their performance. Under the standard setting (checkpointing Attention, LayerNorm, and MoE Experts), our MXFP4 method achieves parity with the FP8 baseline in throughput ($\approx$ 1156 TGS) but reduces memory footprint by 14.8\% . Crucially, this memory headroom allows us to reduce the recomputation scope. When we relax the recomputation strategy to only checkpoint the MLA Up-Projection, both BF16 and FP8 baselines encounter Out-of-Memory (OOM) errors. In contrast, our system operates efficiently within the memory limit (70.11\%), achieving a peak throughput of 1302 TGS. This represents a 12.5\% speedup over the best feasible FP8 configuration and a 16.0\% speedup over the BF16 baseline.  

\textbf{Impact of Recomputation Granularity.}
As shown in Table~\ref{tab:main_results}, the 671B model is primarily constrained by the memory wall. Under FP8, fitting the model requires full recomputation of expert layers, introducing substantial redundant computation. By quantizing activations in the MLP and shared MoE expert layers to MXFP4, we reduce peak memory usage by approximately 15\%, which allows us to skip recomputation for these layers. Despite a small quantization overhead, eliminating backward-pass recomputation yields a net throughput gain of 146 TGS (from 1156 to 1302).

\begin{table}[!htbp]
\centering
\resizebox{\linewidth}{!}{
\begin{tabular}{lcccccc}
\toprule
\textbf{Recomputation Scope} & \multicolumn{2}{c}{\textbf{Baseline (BF16)}} & \multicolumn{2}{c}{\textbf{Baseline (FP8)}} & \multicolumn{2}{c}{\textbf{Ours (MXFP4)}} \\
\cmidrule(lr){2-3} \cmidrule(lr){4-5} \cmidrule(lr){6-7}
 & TGS & Mem& TGS & Mem& \textbf{TGS} & \textbf{Mem}\\
\midrule
\textit{Attn + LN + MoE Experts} & 1122 & 68.29\%& 1157 & 74.20\%& 1156 & \textbf{59.40\%}\\
\textit{Attn + LN + MLP} & OOM & - & OOM & - & \textbf{1248} & 60.62\%\\
\textit{MLA Up Proj Only} & OOM & - & OOM & - & \textbf{1302} & 70.11\%\\
\bottomrule
\end{tabular}
}
\caption{Performance on 671B MoE Model.}
\label{tab:main_results}
\end{table}

We conduct additional experiments on a 236B model in table~\ref{tab:236B_main_results} to enable a controlled comparison across numerical precisions under identical recomputation policies, which is not feasible at the 671B scale due to frequent OOM failures of BF16 and FP8. By fixing the recomputation strategy and varying only precision, we show that MXFP4 reduces peak memory usage by 6.9\% compared to blockwise FP8 when FP4 is applied to the MLP and shared experts, and by 7.2\% when FP4 is extended to the entire MoE, corresponding to an 11\% reduction relative to BF16. Under the most aggressive recomputation regime, MXFP4 consistently outperforms FP8 with an average 7.5\% memory reduction, while BF16 still fails to fit into device memory. Together with the 671B results, these findings confirm that applying FP4 to both MLP and MoE components yields substantial and scalable memory benefits.

\begin{table}[!htbp]
\centering
\resizebox{\columnwidth}{!}{
\begin{tabular}{lcccccc}
\toprule
\textbf{Recomputation Scope}
& \multicolumn{2}{c}{\textbf{Baseline (BF16)}}
& \multicolumn{2}{c}{\textbf{Baseline (FP8)}}
& \multicolumn{2}{c}{\textbf{Ours (MXFP4)}} \\
\cmidrule(lr){2-3}
\cmidrule(lr){4-5}
\cmidrule(lr){6-7}
& TGS & Mem
& TGS & Mem
& \textbf{TGS} & \textbf{Mem} \\
\midrule
\textit{Attn + LN + MoE Experts}
& 1725.16 & 53.22\%
& 1621.96 & 56.81\%
& 1608.68 & \textbf{49.96\%} \\
\textit{Attn + LN + MLP}
& 1838.56 & 60.80\%
& 1746.20 & 57.03\%
& 1616.90 & \textbf{49.85\%} \\
\textit{MLP Up Proj Only}
& OOM & -
& 1831.20 & 66.13\%
& 1722.98 & 58.62\% \\
\bottomrule
\end{tabular}
}
\caption{Performance on 236B MoE Model.}
\label{tab:236B_main_results}
\end{table}


\subsection{Analysis and Ablation Studies}
To understand the source of our performance gains, we benchmark our custom quantization kernels against the standard Transformer Engine (TE) FP8 baselines. 

\begin{figure}[htbp]
    \centering
    \includegraphics[width=0.9\linewidth]{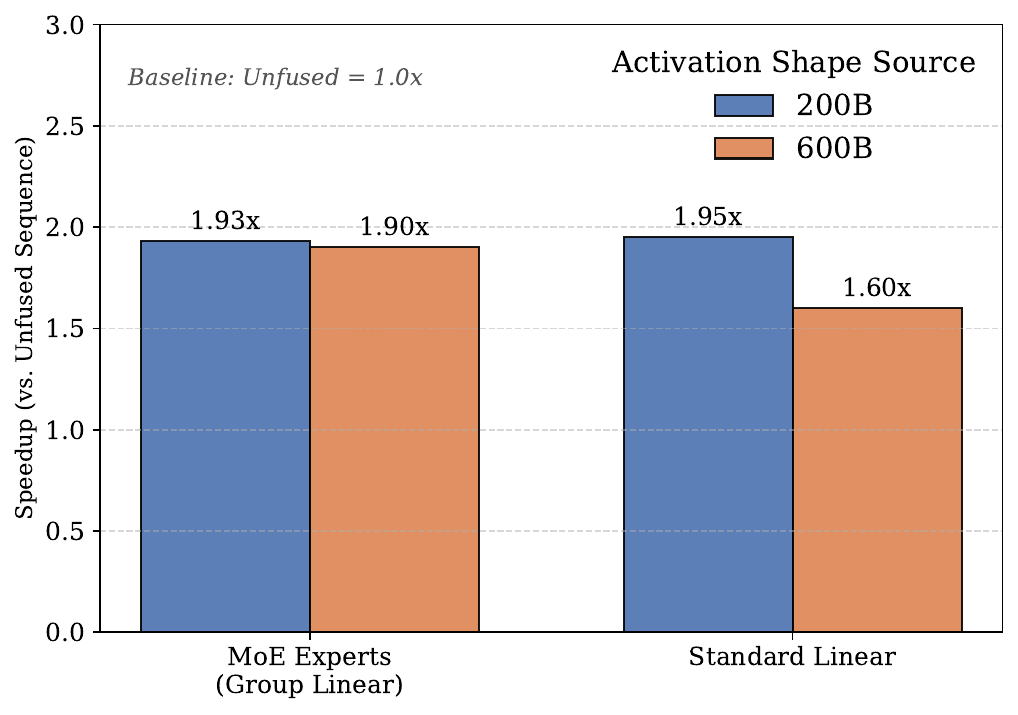}
    \caption{\textbf{Fusion Efficiency (Fused Kernel vs. Dequant + TE Quant).} 
        Our fused approach achieves consistent speedups (1.6x--1.9x) by eliminating intermediate memory writes.}
    \label{fig:kernel_fusion}
\end{figure}

\textbf{Fusion Efficiency} 
First, we validate our kernel fusion strategy in Figure~\ref{fig:kernel_fusion}. Evaluated across tensor shapes representative of 236B and 671B training configurations, our approach—which fuses dequantization, data type casting, and layout transformation into a single kernel—achieves a 1.6x–1.9x speedup compared to a naive sequence of separate operations.

\begin{figure}[htb]
    \centering
    \includegraphics[width=0.9\linewidth]{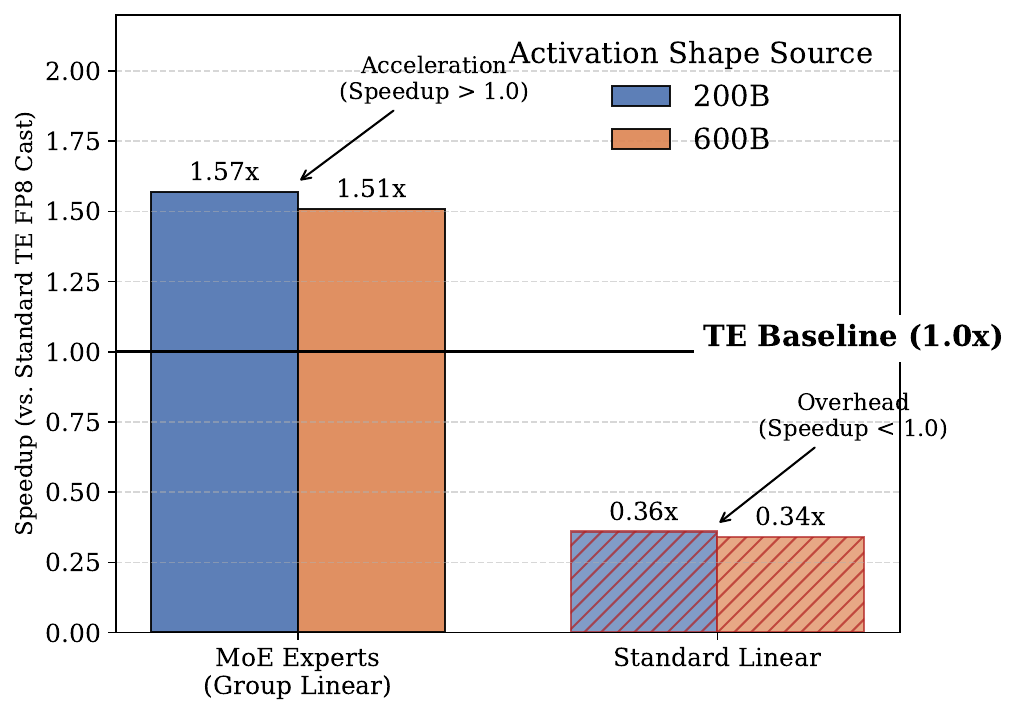}
    \caption{\textbf{Total Quantization Overhead (Ours Full Pipeline vs. TE Baseline).}
        While standard linear layers incur overhead ($<$1.0x), specialized MoE kernels achieve net acceleration ($>$1.5x).}
    \label{fig:e2e_overhead}
\end{figure}

\textbf{End-to-End Kernel Performance}
When comparing our total operator latency (Quantization + Fused Dequantization) against the TE FP8 baseline in Figure~\ref{fig:e2e_overhead}, we observe two distinct behaviors:

\begin{itemize}
\item Standard Linear Layers: For standard dense projections (e.g., in Attention blocks), the additional overhead of on-the-fly quantization results in a 0.33x relative speedup (i.e., higher latency) compared to the highly optimized TE baseline.
\item MoE Expert Layers (Group Linear Layers): Crucially, for the computation-heavy MoE expert layers, our specialized msplit kernel significantly outperforms the Transformer Engine FP8 split quantize baseline, achieving a 1.43x–1.53x speedup. This suggests that our custom implementation handles the grouped memory access patterns of MoE experts more efficiently than the generic library calls.
\end{itemize}
Since the MoE expert computation dominates the total FLOPs of the 671B model, this acceleration in the critical path, combined with the reduced recomputation, successfully amortizes the overhead incurred in the standard linear layers, leading to the overall system throughput improvement.

\subsection{Convergence Verification}
To evaluate the training stability of our MXFP4 quantization scheme, we reproduce a configuration consistent with recent large-scale LLM training recipes and the official MXFP4 training methodology. All models are trained using AdamW with a cosine learning rate decay on a 16B-parameter model. We fix the global batch size to 4800 sequences and a context length of 4096 tokens, matching the conditions used in our primary comparisons. The optimizer uses $\beta_1=0.9$, $\beta_2=0.95$, and weight decay of $1.0\times10^{-1}$; gradients are clipped to a global norm of 1.0. We apply the same weight initialization, dropout, and tokenizer settings across BF16 and MXFP4 runs to prevent confounding factors. Both BF16 and MXFP4 experiments are run on identical hardware clusters using the same mixed-precision and tensor-parallel infrastructure.

\begin{figure}[htbp]
    \centering
    \includegraphics[width=\linewidth]{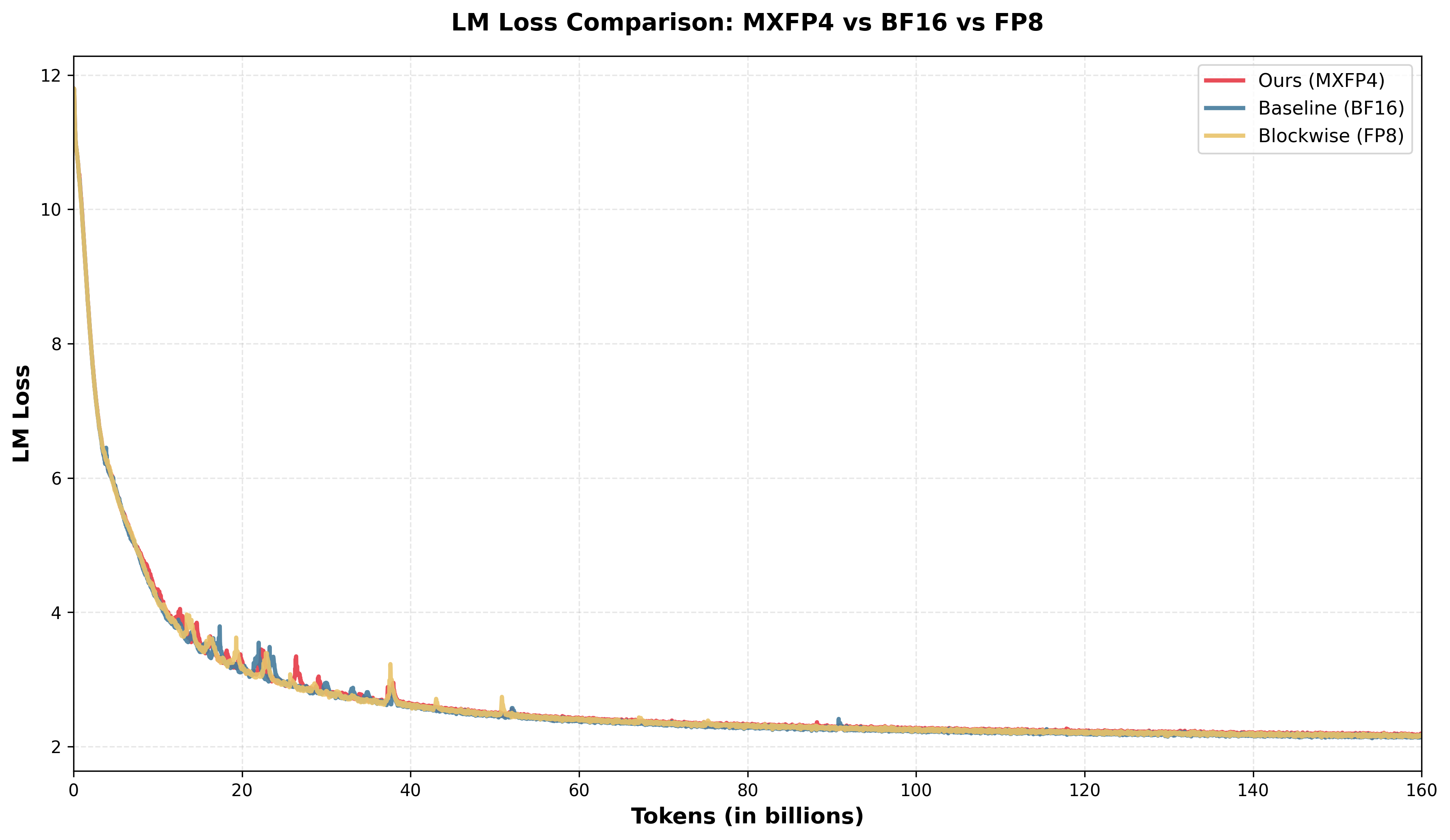}
    \caption{LM Loss Comparison.}
    \label{fig:loss}
\end{figure}

As illustrated in Figure~\ref{fig:loss}, the loss trajectory of our method strictly follows the BF16 baseline, with no observed instability or divergence. Loss trajectories and quantization-induced deviations are evaluated after consuming 160B training tokens. Quantitatively, FP8 and MXFP4 incur relative loss deviations of $+0.29\%$ and $+0.61\%$ with respect to BF16, respectively, indicating that MXFP4 maintains stable optimization despite more aggressive quantization.

%% file: sections/05-conclusion.tex
\section{Conclusion}
\label{sec:conclusion}


We present a software-emulated MXFP4 training framework that enables FP4-level activation and communication efficiency on Hopper GPUs without native 4-bit Tensor Core support. Key enablers include direct FP4-to-FP8 bitwise conversion, scaling-aware layout transformations, and fused CUDA kernels that minimize casting overhead and memory traffic. On a 671B MoE model, our method reduces peak memory usage, increases training throughput, and achieves convergence comparable to FP8. Overall, these results demonstrate that FP4 efficiency can be practically realized on widely deployed Hopper hardware.

%% file: sections/appendix-a.tex
\section{Appendix: Code and Implementation Notes}
\label{sec:appendix-code}

We provide the full implementation and system integration instructions at \href{https://github.com/anonymous-git-2026/MXFP4-Hopper}{github}. 
The codebase enables MXFP4-based training on Hopper-class GPUs and includes:

\begin{itemize}
\item \textbf{\texttt{Custom Kernels}}: Low-level CUDA implementations for MXFP4 quantization, dequantization, and format conversion.
\item \textbf{\texttt{Framework Modifications}}: Patches for Megatron-LM (v0.15.0rc7), NVIDIA Transformer Engine (v2.8), and DeepEP (v1.2.1) to support FP4-aware training and communication.
\item \textbf{\texttt{Usage Guide}}: Build and integration instructions to reproduce the results in the main paper.
\end{itemize}

Please refer to the supplementary README file for a detailed breakdown of repository structure and usage.